\documentclass[11pt]{article}

\usepackage[final]{acl}

\usepackage{times}
\usepackage{latexsym}

\usepackage[T1]{fontenc}
\usepackage[utf8]{inputenc}

\usepackage{microtype}

\usepackage{inconsolata}

\usepackage{graphicx}
\usepackage{hyperref}
\usepackage{twemojis}
\usepackage{booktabs}
\usepackage{fontspec}
\usepackage{array}
\usepackage{tablefootnote}

\usepackage{tabularray}
\usepackage{cellspace}
\newcommand\cincludegraphics[2][]{\raisebox{-0.3\height}{\includegraphics[#1]{#2}}}
\usepackage{setspace}

\newcommand{\BCE}{{\small BCE}}

\newfontfamily{\cuneiformfont}{NotoSansCuneiform}[Extension = .ttf, UprightFont = *-Regular]
\newcommand{\cuneiform}[1]{{\cuneiformfont #1}}

\newcommand{\AN}{\cuneiform{\char"1202D\relax}}

\newcommand{\EN}{\cuneiform{\char"12097\relax}}
\newcommand{\KID}{\cuneiform{\char"121A4\relax}}
\newcommand{\LUGAL}{\cuneiform{\char"12217\relax}}
\newcommand{\KUR}{\cuneiform{\char"121B3\relax}}
\newcommand{\RA}{\cuneiform{\char"1228F\relax}}
\newcommand{\AB}{\cuneiform{\char"1200A\relax}}
\newcommand{\BA}{\cuneiform{\char"12040\relax}}
\newcommand{\URU}{\cuneiform{\char"12337\relax}}
\newcommand{\NE}{\cuneiform{\char"12248\relax}}
\newcommand{\KA}{\cuneiform{\char"12157\relax}}
\newcommand{\GI}{\cuneiform{\char"120B5\relax}}
\newcommand{\NA}{\cuneiform{\char"1223E\relax}}
\newcommand{\NI}{\cuneiform{\char"1224C\relax}}
\newcommand{\TA}{\cuneiform{\char"122EB\relax}}

\newcommand{\BU}{\cuneiform{\char"1204D\relax}}

\title{\textit{SumTablets} \texttwemoji{amphora}:\\A Transliteration Dataset of Sumerian Tablets}

\author{Cole Simmons \\
  Stanford University \\
  \texttt{coles@stanford.edu} \\\And
  Richard Diehl Martinez \\
  University of Cambridge \\
  \texttt{rd654@cam.ac.uk} \\\And 
  Dan Jurafsky \\
  Stanford University \\
  \texttt{jurafsky@stanford.edu} \\}

\begin{document}
\maketitle
\begin{abstract}

Sumerian transliteration is a conventional system for representing a scholar's interpretation of a tablet in the Latin script. Thanks to visionary digital Assyriology projects such as ETCSL, CDLI, and Oracc, a large number of Sumerian transliterations have been published online, and these data are well-structured for a variety of search and analysis tasks. However, the absence of a comprehensive, accessible dataset pairing transliterations with a digital representation of the tablet's cuneiform glyphs has prevented the application of modern Natural Language Processing (NLP) methods to the task of Sumerian transliteration.

To address this gap, we present \textit{SumTablets}, a dataset pairing Unicode representations of \textbf{91,606 Sumerian cuneiform tablets} (totaling \textbf{6,970,407 glyphs}) with the associated transliterations published by Oracc. We construct \textit{SumTablets} by first preprocessing and standardizing the Oracc transliterations before mapping each reading back to the Unicode representation of the source glyph. Further, we retain parallel structural information (e.g., surfaces, newlines, broken segments) through the use of special tokens. We release \textit{SumTablets} as a Hugging Face Dataset (CC BY 4.0) and open source data preparation code via GitHub.

Additionally, we leverage \textit{SumTablets} to implement and evaluate two transliteration baselines: (1) weighted sampling from a glyph's possible readings, and (2) fine-tuning an autoregressive language model. Our fine-tuned language model achieves an average transliteration character-level F-score (chrF) of 97.55, demonstrating the immediate potential of transformer-based transliteration models in allowing experts to rapidly verify generated transliterations rather than manually transliterating tablets one-by-one.

\end{abstract}

\begin{tblr}{colspec = {Q[c,m]|X[l,m]}, stretch = 0}
    \cincludegraphics[width=1.4em, keepaspectratio]{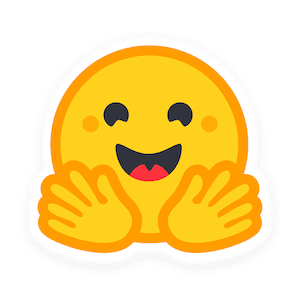} & {\footnotesize{\href{https://huggingface.co/datasets/colesimmons/SumTablets}{colesimmons/SumTablets}} \tiny{(CC BY 4.0)}} \\
    \cincludegraphics[width=1.2em, keepaspectratio]{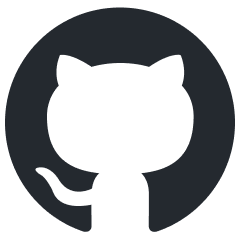} & {\footnotesize{\href{https://github.com/colesimmons/SumTablets}{colesimmons/SumTablets}}}
\end{tblr}

\section{Introduction}

\begin{figure}
    \centering
    \includegraphics[angle=90,origin=c,scale=0.23]{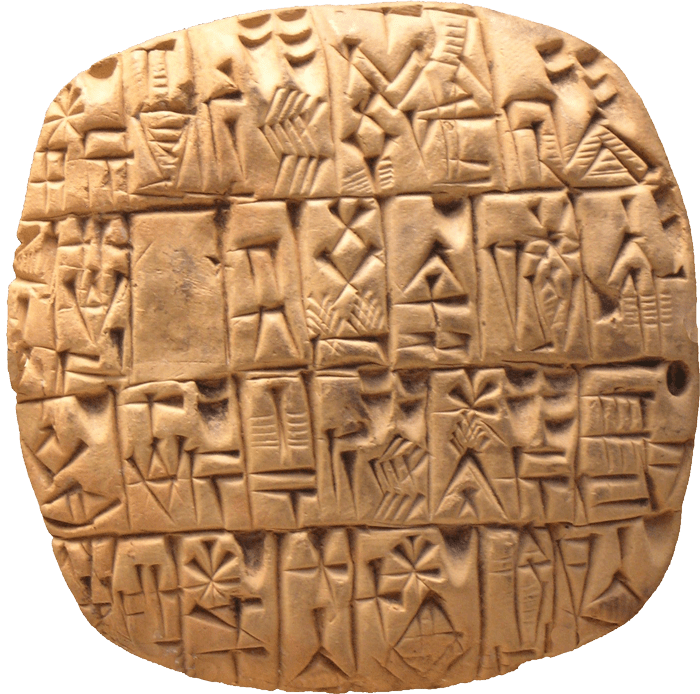}
    \caption{An administrative Sumerian cuneiform tablet from Shuruppak, dated to the Early Dynastic IIIa period (ca. 2500 {\small BCE}). \cite{Cuneiform_Image_BM_15826}}
    \label{fig:sum-tablet}
    \vspace{-1em}
\end{figure}

\begin{table*}[h!]
    \centering  
    \small % Smaller font size
    \begin{tabular}{ p{0.08\textwidth}|p{0.14\textwidth}|p{0.14\textwidth}|p{0.22\textwidth}|p{0.27\textwidth} }
        \toprule
        \textbf{ID} &
        \textbf{Period} & 
        \textbf{Genre} &
        \textbf{Glyphs (Inputs)} &
        \textbf{Transliteration (Targets)} \\
        \midrule
        % Row 1
        Q001103 &
        Early  & 
        Royal &
        {\scriptsize <SURFACE>} &
        {\scriptsize <SURFACE>} \\
        % Row 2
         &
        Dynastic IIIb & 
        Inscription &
        {\AN\EN\KID} &
        \{d\}en-lil\(_2\) \\
        % Row 3
        & & & {\LUGAL\KUR\KUR\RA} & lugal kur-kur-ra \\
        % Row 4
        & & & {\AB\BA\AN\AN\URU\NE\KID} & ab-ba dingir-dingir-re\(_2\)-ne-ke\(_4\) \\
        % Row 5
        & & & {\KA\GI\NA\NI\TA} & inim gi-na-ni-ta \\
        % Row 6
        & & & \dots & \dots \\
        \bottomrule
    \end{tabular}
    \vspace{0.5em}
    \caption{A sample paired glyph--transliteration example from \textit{SumTablets}, dating ca. 2600--2300 \BCE.}
    \label{fig:dataset-example}
\end{table*}

% ==== Who cares? ====
Sumerian is the world's earliest attested written language, marking the transition from prehistory into history as well as reflecting a rich written tradition spanning three thousand years. These texts are an invaluable resource in the study of ancient Near Eastern culture, politics, economics, and more.

During the latter half of the fourth millennium {\small BCE}, a sophisticated record-keeping system emerged in southern Mesopotamia, now known as proto-cuneiform \cite{Selz_2020}. Over time this system evolved\footnote{There continues to be considerable ambiguity and disagreement about the extent to which evolution occurred gradually or was the result of a single inventor. For a more comprehensive treatment of the topic, see \cite{Sproat_2023}.} to handle natural language. By about 2900 {\small BCE} this writing system, known as \textit{cuneiform}, is concretely recognizable as encoding Sumerian.

% ==== Unicode ====

Mesopotamian scribes originally devised the cuneiform script to write Sumerian. This script was later adapted to encode other languages throughout the Near East, such as Akkadian. To form glyphs, scribes would typically compose stylus impressions on a wet clay tablet\footnote{Although not all texts are clay or in the form of a tablet, we follow Assyriological convention by referring to texts generically as tablets.}. Because cuneiform writing was impressed or inscribed on durable materials, texts have survived to the present in tremendous quantity \citep{Finkel_Taylor_2015}. Uncovered during archaeological excavations of ancient cities beginning in the nineteenth century {\small CE}, these tablets had to be subsequently deciphered. Deciphering Sumerian, a language isolate, proved particularly challenging, and some periods and genres are still not completely understood.

Sumerian cuneiform glyphs are frequently polyvalent; that is, they have many possible readings (of no necessary semantic or phonetic relation) depending on the context. For instance, \cuneiform{\KA} can be read as \textit{ka} ``mouth,'' $\textit{dug}_{4}$ ``to speak,'' $\textit{kiri}_{3}$ ``nose,'' \textit{zuh} ``to steal,'' the syllable \textit{ka}, and more. When reading a tablet, an Assyriologist must often consider various possibilities for each glyph to achieve a set of consistent readings. They represent their interpretation through the process of transliteration.

% Background to Transliteration
Transliteration is a modern, conventional system for representing Sumerian in the Latin alphabet. Conventions were established at various points in the modern, 150-year history of Sumerian decipherment and do not necessarily reflect the current understanding of Sumerian phonology or morphology. In transliterations, homophones are distinguished via subscripts; for instance, $e$ and $e_2$ are homophonic---but semantically unrelated---readings of different glyphs. Additionally, hyphens are used to join nominal/verbal roots with affixes \citep{Michalowski_2004}.

In 1996, the Electronic Text Corpus of Sumerian Literature (ETCSL) \cite{ETCSL} project began publishing transliterations online. This project became archival in 2006, soon followed by other projects such as the Cuneiform Digital Library Initiative (CDLI) \cite{CDLI} and the Open Richly Annotated Cuneiform Corpus (Oracc) \cite{ORACC}. Thanks to these and other projects, a large number of transliterations have been published online and their data made available for use with open licenses. Our work would not be possible without the decades of dedicated efforts by contributors to these projects.

Because Assyriologists are reading from either the physical text or an image, no digital representation of the original text's glyphs is typically recorded. Today, most cuneiform glyphs have been added to Unicode\footnote{All online Sumerian data aggregation and collaboration was limited to ASCII for more than a decade: The first cuneiform was added to Unicode in 2006.}. However, easily accessible\footnote{We define \textit{easily accessible} as being easily utilized programmatically and requiring no or minimal Assyriological expertise to contribute to development of models based on these datasets.}, standardized datasets of paired Sumerian Unicode glyphs and transliterations remain limited, barring the development of transliteration models.

% Text cross-lingual approaches
In this paper, we present the first large-scale, easily accessible dataset of \textbf{91,606 Sumerian tablets} as glyph--transliteration pairs, containing a total of \textbf{6,970,407 glyphs}. We additionally include IDs\footnote{IDs are consistent with those in Oracc and CDLI.}, period, and genre metadata for each tablet to be used for results analysis.

Our dataset, \textit{SumTablets}, is derived from a collection of publicly available Sumerian language resources, primarily the Electronic Pennsylvania Sumerian Dictionary (ePSD2) \cite{ePSD2} and the Oracc Sign List (OSL) \cite{OSL}. These projects aggregate and index transliteration data from across Oracc, which shares data with CDLI and includes data from other current and former projects\footnote{\href{https://oracc.museum.upenn.edu/epsd2/credits/index.html}{ePSD2 credits}}.

Because of how they are formatted and because they do not include parallel Unicode glyph tablet representations, however, the data on Oracc are not immediately suited for glyph-to-transliteration tasks. We preprocess these data to clean and standardize them, converting structure-related annotations into special tokens. Then, since a given reading maps back to only one glyph, we utilize Unicode--reading dictionaries provided by ePSD2 and OSL to convert each reading back into its source glyph. 

We upload our dataset to Hugging Face \cite{HuggingFace}, the largest and most widely utilized library for sharing datasets for machine learning tasks. We intend to use Hugging Face's git-based version control to provide experiment reproducibility over time, with versions containing snapshots of the continuously updated Oracc data.

Our dataset, \textit{SumTablets}, builds on previous open-source projects by:
\begin{enumerate}
    \item \textbf{being the largest dataset of parallel glyph--transliteration examples.}
    \item \textbf{standardizing the data available in Oracc}, optimizing formatting for the transliteration task while maintaining the morphosyntactic fidelity of the texts.
    \item \textbf{vastly facilitating the use of this data in machine learning projects}, simplifying access via the common Hugging Face Datasets library.
\end{enumerate}

% ==== Models ====

% ==== Evaluation ====
Using our dataset, we develop and compare two baseline transliteration approaches. The first is a weighted dictionary mapping; for each glyph we sample one of the glyph's possible readings according to its frequency. The second is a language model that we fine-tune for the glyph-to-transliteration task. As far as we are aware, we are the first to develop an automatic Sumerian transliteration model. Evaluated on a held-out test set, the dictionary-lookup approach obtains a character-level F-score (chrF) \citep{Popovic_2015} of 61.22, while the fine-tuned model achieves a chrF score of 97.54.

% ==== Analysis / Limitations ====
Our goals in releasing this dataset are to facilitate the development of transliteration models and to demonstrate the potential of adapting large pretrained multilingual models for the task. We envision web-based tooling built on top of neural transliteration models helping Assyriologists to generate transliterations more quickly---allowing them to rapidly validate model outputs rather transliterating each tablet from scratch---and target review of potential errors in existing transliterations. Additionally, transliteration models serve as an essential step in eventually developing a complete Sumerian translation pipeline. Finally, as a language isolate, Sumerian poses a unique syntactic challenge for cross-lingual models, and opens new avenues of research into the transfer of language understanding. 

% =============================================
% ============== RELATED WORK =================
% =============================================
\section{Related Work}
\label{section:related-work}

% ==== SOURCE DATA ====

% ==== SUMERIAN ====
To the best of our knowledge, our work represents the first to formulate Sumerian transliteration as an NLP task and to develop a transliteration model. However, prior works have utilized NLP techniques for other tasks in parsing and analyzing Sumerian cuneiform. The Machine Translation and Automated Analysis of Cuneiform Languages (MTAAC) project \citep{Page-Perron_EtAl_2017} sought to develop a pipeline for Sumerian annotation, translation, and information extraction, working primarily with Ur III transliterations. \citeauthor{Chiarcos_EtAl_2018} expanded this data to include the Electronic Text Corpus of Sumerian Royal Inscriptions (ETCSRI) \citep{Zolyomi_EtAl_2019}. \citeauthor{Bansal_EtAl_2021} then used MTAAC data in conjunction with CDLI and ETCSL data to train models for part-of-speech (POS) tagging, named entity recognition (NER), and translation, aiming primarily to build generalizable cross-lingual methods for performing these tasks on low-resource languages. The COMPASS \cite{Veldhuis_2024} also explores using cuneiform data for research tasks, such as reconstructing social graphs. Perhaps most similar to our work, \citeauthor{Gordin_EtAl_2020} develop a neural network to automatically transliterate Akkadian from Unicode cuneiform glyphs.

% ==== DATASETS ====
Others have built datasets also representing tablets' glyphs in Unicode. \citeauthor{Jauhiainen_EtAl_2019} utilized Oracc dataset to build a dataset of 13,662 tablets for the task of language and dialect identification. More recently, \citeauthor{Chen_EtAl_2023} used CDLI data to create CuneiML, a dataset of 38,947 tablets with photos, Unicode glyphs, transliterations, and metadata, also designed primarily for classification tasks. Both of these datasets include both Sumerian and Akkadian texts, whereas our dataset only includes monolingual Sumerian texts. Furthermore, our dataset is larger, designed specifically for the transliteration task, and is easily accessible through Hugging Face.

% ==== SUMERIAN ====
Outside of NLP, an exciting area of research is using computer vision methods to identify cuneiform signs from images \citep{Dencker_EtAl_2020}. Efforts in visual classification and transcription of cuneiform are enabled by projects that have open-sourced high-quality 2D and 3D images of tablets \citep{Dahl_EtAl_2019, Mara_Homburg_2023}. And beyond cuneiform, \citeauthor{Assael_EtAl_2022} used deep learning methods to restore fragmented ancient texts in ancient Greek.

% ==== MODELS / BENCHMARKS ====
As Sumerian is a low-resource language, it is infeasible to train a transformer-based language model on Sumerian from scratch rather than adapting cross-lingual representations in existing models. Fortunately, the recent success of large cross-lingual NLP models such as mBERT \citep{Devlin_EtAl_2019}, XLM-R \cite{Conneau_EtAl_2020}, m-T5 \cite{Liu_EtAl_2020}, and BLOOM \citep{BigScienceWorkshop_EtAl_2023} have steadily raised the bar for zero- and few-shot cross-lingual performance on benchmarks such as XTREME \citep{Hu_EtAl_2020} and MEGA \citep{Ahuja_EtAl_2023}. Recently, benchmarks to measure a model's ability to perform NLP tasks in extremely low-resource and orthographically-diverse languages have emerged, such as IndicXNLI \citep{Aggarwal_EtAl_2022} for low-resource Indian languages, and \citeauthor{Sukhareva_EtAl_2017} who develop a POS tagging benchmark for Hittite, another cuneiform language. \textit{SumTablets} marks the first benchmark for Sumerian neural machine transliteration.

% =============================================
% ============== CREATING SUMTABLETS  =========
% =============================================
\section{Creating \textit{SumTablets} \texttwemoji{amphora} }

\textit{SumTablets} is built upon the metadata and transliterations provided by ePSD2 via JSON files\footnote{https://oracc.museum.upenn.edu/epsd2/json}. These transliterations were created or manually typed by scholars working in different projects around the world over decades of evolving knowledge of Sumerian vocabulary and grammar; they also contain extensive (but not useful for our purposes) embedded ASCII annotation. We begin by preprocessing the transliterations to normalize conventions, remove annotations, and convert formatting information into special tokens. Then, we use dictionaries built from ePSD2 and OSL resources to map each reading back to a Unicode representation of its source glyph. The result is a set of Unicode glyph--transliteration pairs with parallel formatting, allowing language models to most effectively learn the relationships between the two representations.

\subsection{Initial Data Cleaning}

We first parse and type-check the ePSD2 JSON data using custom Pydantic\footnote{https://docs.pydantic.dev/latest/} classes. The transliterations are structured in a recursive format called cdl (for the three node types: chunk, delimiter, and lemma) at the document level, which we navigate in order to reconstruct the transliteration as a single string with embedded formatting information.

We then remove annotations embedded in the transliterations. Many of these represent the editor's interpretation beyond what is visible on the tablet; for instance, text enclosed in square brackets represents the editor's belief of what was originally in a now-missing segment. While this information is academically useful, it can inject an undesirable bias when training transliteration models. Our goal is to best represent only what is on the tablet. We remove text enclosed in square brackets (broken) and single angle brackets (graphemes must be supplied for the sense but are not present), replacing the former with a \textbf{...} special token to indicate breakage. For text enclosed in upper square brackets (partially visible) and double angle brackets (graphemes are present but most be excised for the sense), we remove the notation but retain the text. These examples are a few of many conventions are used in the provided transliterations. For each type, we either remove the notation but retain the text, remove the notation and the text, or replace the notation and text with a special token (described in subsection \ref{subsec:extra-semantic-tokens}).

The Oracc data are supplied with metadata that varies depending on the project in which a tablet was digitized. After performing an inner join on all of the data, we found the period and genre to be the most salient, universally-supplied metadata; because we provide the original Oracc IDs, removed fields can easily be reintegrated.

\subsection{Mapping transliterations to glpyhs}

For each of the transliterations, we generate the associated glyphs in three steps: 
\begin{enumerate}
    \item First, we split each transliteration by spaces to get a list of words, which we then split further into individual glyph readings (i.e., morphemes).
    \item Next, for each reading, we look up the corresponding glyph name. Each glyph in Sumerian has a conventional name that is an uppercase version of one of its readings; for instance, the glyph \KA\ is referred to as {\small KA}. Like most glyphs, it can be read a number of different ways (e.g., \textit{ka}, \textit{$dug_4$}, \textit{inim}). Importantly, these readings are readings only of {\small KA} and can be mapped back to it. If we cannot ascertain the glyph name, we replace the reading with {\small <UNK>}. Sign names are often used in place of a reading (to say that the reading is uncertain), in which case we replace the reading with {\small <UNK>} but will still use the corresponding Unicode. The first row of Table \ref{tab:preprocessing-steps} shows the proportion of readings that we are able to map to glyph names. 
    \item Finally, we convert each glyph name to the Unicode representation of that glyph name; for instance, we convert {\small BU} to \BU. For the rare glyphs that are not represented in Unicode, we replace both the glyph and associated reading with {\small <UNK>} tokens. The bottom row of Table \ref{tab:preprocessing-steps} shows the proportion of glyphs names that we able to map to Unicode. 
\end{enumerate} 

To map from transliteration to glyph name and from glyph name to Unicode, we leverage ePSD2 and OSL.

\begin{table}[!h]
    \centering
    \small
    \begin{tabular}{l|l}
        \toprule
        Preprocessing Step & Success Rate \\
        \midrule
        Readings $\rightarrow$ Glyph Name & 6,724,498 (99.93\%) \\
        Glyph Name $\rightarrow$ Unicode & 6,638,081 (99.96\%) \\
        \bottomrule
    \end{tabular}
    \caption{Preprocessing steps with associated amount of maintained glyphs in constructing \textit{SumTablets}. }
    \label{tab:preprocessing-steps}
\end{table}

\subsection{Extra-semantic tokens}
\label{subsec:extra-semantic-tokens}
In addition to the aforementioned preprocessing steps, we add the following special tokens to maintain structural information about each tablet in corresponding locations in the glyph and transliteration examples:

\begin{itemize}
    \item \textbf{<SURFACE>} -- The start of a surface. For a tablet, this may be the start of the obverse or reverse side. For other types of artifacts (like statues), the number of surfaces and their relationship to each other depends on the form.
    % TODO what percent of texts are multi-surface?
    \item \textbf{\textbackslash n} -- A line break. These are important to retain because it is extremely rare that a wordform runs over to a subsequent line.
    \item \vspace*{-0.1em} \textbf{...} -- Breakage. Ellipses on their own line indicate an indeterminate number of missing lines, while ellipses on a line with text indicate an indeterminate number of missing glyphs.
    \item \vspace*{-0.1em} \textbf{<RULING>} -- A horizontal line drawn by the scribe to separate sections of the tablet.
    \item \vspace*{-0.1em} \textbf{<COLUMN>} -- The start of a new column of text. Not all tablets are formatted in columns.
    \item \vspace*{-0.1em} \textbf{<BLANK\_SPACE>} -- The scribe left some amount of blank space before continuing on.
\end{itemize}

\subsection{Metadata}
\label{subsection:meta-data}

As part of the dataset, we include additional metadata associated with each tablet: the time period each tablet dates from and the semantic genre of each tablet (e.g. administrative, legal). In total, we define 10 unique time periods and 14 genres (see Table \ref{tab:period-genre-composition}).

\begin{table}[h!]
    \centering
    \newcolumntype{L}[1]{>{\raggedright\arraybackslash}p{#1}}
    \newcolumntype{R}[1]{>{\raggedleft\arraybackslash}p{#1}}
    \begin{tabular}{L{3.3cm}|R{1.1cm}R{0.8cm}R{0.8cm}}
        \toprule
            \textbf{Period}  & Train & Val & Test \\
        \midrule
            Ur III &  71,116 & 3,951 & 3,951 \\
            Old Akkadian & 4,766 & 265 & 265 \\
            Early Dynastic IIIb  & 3,467 & 192 & 192 \\
            Old Babylonian  &  1,374 & 73 & 73 \\
            Lagash II  &  788 & 44 & 44 \\
            Early Dynastic IIIa  &  755 & 42 & 42 \\
            Early Dynastic I-II &  77 & 4 & 4 \\
            Unknown &   68 & 4 & 4 \\
            Neo-Assyrian  &  20 & 1 & 1 \\
            Neo-Babylonian  &  14 & 1 & 1 \\
            Middle Babylonian &  7 & 0 & 0 \\
        \midrule
        \textbf{Total} & 82,452 & 4,577 & 4,577 \\
        \midrule
        \midrule
     \end{tabular}
     \begin{tabular}{L{3.3cm}|R{1.1cm}R{0.8cm}R{0.8cm}}
            \textbf{Genre} & Train & Val & Test \\
        \midrule
            Administrative & 77,193 & 4,259 & 4,291 \\
            Royal Inscription & 2,611 & 151 & 146 \\
            Literary & 1,000 & 63 & 62 \\
            Letter & 718 & 48 & 33 \\
            Legal & 544 & 35 & 36 \\
            Unknown & 269 & 14 & 7 \\
            Lexical & 69 & 0 & 0 \\
            Liturgy & 40 & 4 & 1 \\
            Math/Science & 8 & 3 & 1 \\
        \midrule
        \textbf{Total} & 82,452 & 4,577 & 4,577 \\
        \bottomrule
    \end{tabular}
    \caption{Composition of tablets by period and genre in \textit{SumTablets}.}
    \label{tab:period-genre-composition}
\end{table}

\subsection{Data Partitions}
\label{subsection:data-partition}

For the purposes of developing automatic transliteration approaches, we split our corpus into train, validation, and test partitions using a 90\%/5\%/5\% split. As an artifact both of what was produced as well as what sites have been excavated, there is a considerable imbalance in the number of examples between historical periods and genres. To ensure that we are training, validating, and testing evenly on how the language was used over time, we stratify the splits by period---Table \ref{tab:period-genre-composition} shows the number of examples in each by split. Because the genres of texts produced correlates strongly with period, stratifying by period results in a nearly equal split of genres, also shown in Table \ref{tab:period-genre-composition}. Importantly, we removed the lexical texts before splitting, and then added them back to the train set after.\footnote{Lexical texts are lists of words that were used in scribal training. We believe that it does not make sense to evaluate against them, but leave it up to the user to decide whether they provide productive noise during training.}

\begin{table*}[h!]
    \centering
\begin{tabular}{lrr|lrr}
\toprule
\multicolumn{3}{c|}{\textbf{Period}} & \multicolumn{3}{c}{\textbf{Genre}} \\
\midrule
Category & Dictionary & Neural & Category & Dictionary & Neural \\
\midrule
Ur III & 62.89 & \textbf{98.46} &
Administrative & 63.15 & \textbf{98.14} \\
Old Akkadian & 64.52 & \textbf{94.03} &
Royal Inscription & 54.58 & \textbf{95.15} \\
Early Dynastic IIIb & 62.51 & \textbf{97.08} &
Literary & 37.73 & \textbf{90.67} \\
Old Babylonian & 37.70 & \textbf{90.38} &
Letter & 47.43 & \textbf{90.99} \\
Lagash II & 58.55 & \textbf{93.97} &
Legal & 56.19 & \textbf{96.14} \\
Early Dynastic IIIa & 67.85 & \textbf{95.02} &
Unknown & 69.84 & \textbf{97.58} \\
Early Dynastic I-II & 73.72 & \textbf{96.82} &
Liturgy & 55.92 & \textbf{77.68} \\
Unknown & 64.98 & \textbf{89.87} &
Math/Science & 62.00 & \textbf{95.12} \\
Neo-Assyrian & 40.83 & \textbf{89.79} & & & \\
Neo-Babylonian & 42.47 & \textbf{97.81} & & & \\
\midrule
\textbf{Overall} & 61.22 & \textbf{97.54} & & 61.22 & \textbf{97.54} \\
\bottomrule
\end{tabular}

    \caption{Results by period and genre. Average chrF scores of transliterations generated in the dictionary baseline compared against those generated  in the neural baseline.}
    \label{tab:results}
\end{table*}

% =============================================
% ============== EVALUATION ===================
% =============================================
\section{Evaluating Transliteration Performance}

The scale and standardization of \textit{SumTablets} enables new methods to be applied to the task of Sumerian transliteration. In this section, we leverage our dataset to develop and compare two transliteration approaches: a straight-forward `dictionary baseline' and a `neural baseline'. First, we define the transliteration task.

\subsection{The Transliteration Task}
\label{subsection:transliteration-task-definition}

We model transliteration as a sequence-to-sequence conversion task, where the input sequence is defined as glyphs and the output as a sequence of alpha-numeric characters, hyphens and white spaces. Table \ref{fig:dataset-example} illustrates example pairs of input (glyphs) and output (transliterations). As we model it, the transliteration task is more akin to a translation task, where each input sequence can be mapped to a large space of output sequences, rather than a token classification task. Given our framing of the transliteration task, we use character-level chrF score as the evaluation metric, defined as:

\begin{equation}
    \texttt{chrF} = (1+\beta)^2 \frac{\texttt{chrP} \cdot \texttt{chrR}}{\beta^2 \cdot \texttt{chrP} + \texttt{chrR}}
\end{equation}

where $\texttt{chrP}$ and $\texttt{chrR}$ stand in for character-level precision and recall scores. Throughout our analysis, we set $\beta=2$, and use a character n-gram order of 6, as proposed by \citeauthor{Popovic_2015}. We compute the chrF score over the transliterated tokens for each tablet individually and then average these scores together over the dataset.

\subsection{Dictionary Baseline}
\label{subsection:manual-baseline}

As part of previous transliteration efforts, Sumerian language experts have hand-crafted dictionaries that map a glyph to all possible readings of that glyph. We cross-analyze our dataset with the ePSD2 and OSL Sumerian dictionaries and find that the average number of different readings for a glyph, weighted by glyph frequency, is 22.17. 

The availability of these dictionaries yields a simple automatic Sumerian transliteration approach: for each glyph in the test set, sample over its possible readings in proportion to their frequency\footnote{We recorded occurrence counts in the process of constructing the dataset.}. This baseline results in an average chrF score of 61.22. 

\subsection{Neural Baseline}
\label{subsection:neural-baseline}

We explore whether the cross-lingual abilities of existing multilingual language models can be leveraged to solve the Sumerian transliteration task. Although Sumerian is a language isolate, it shares grammatical features with other modern languages: like Basque, it has ergative--absolutive alignment; like Turkish and Japanese, it is agglutinative; and like Korean, it is SOV \citep{Michalowski_2004}. Therefore, the key to our approach is to leverage XLM-R \citep{Conneau_EtAl_2020}, a transformer language model pre-trained on over 100 languages.

\begin{figure}[ht!]
    \centering
    \vspace*{-1cm}
    \hspace*{-0.24\linewidth}
    \includegraphics[width=1.4\linewidth]{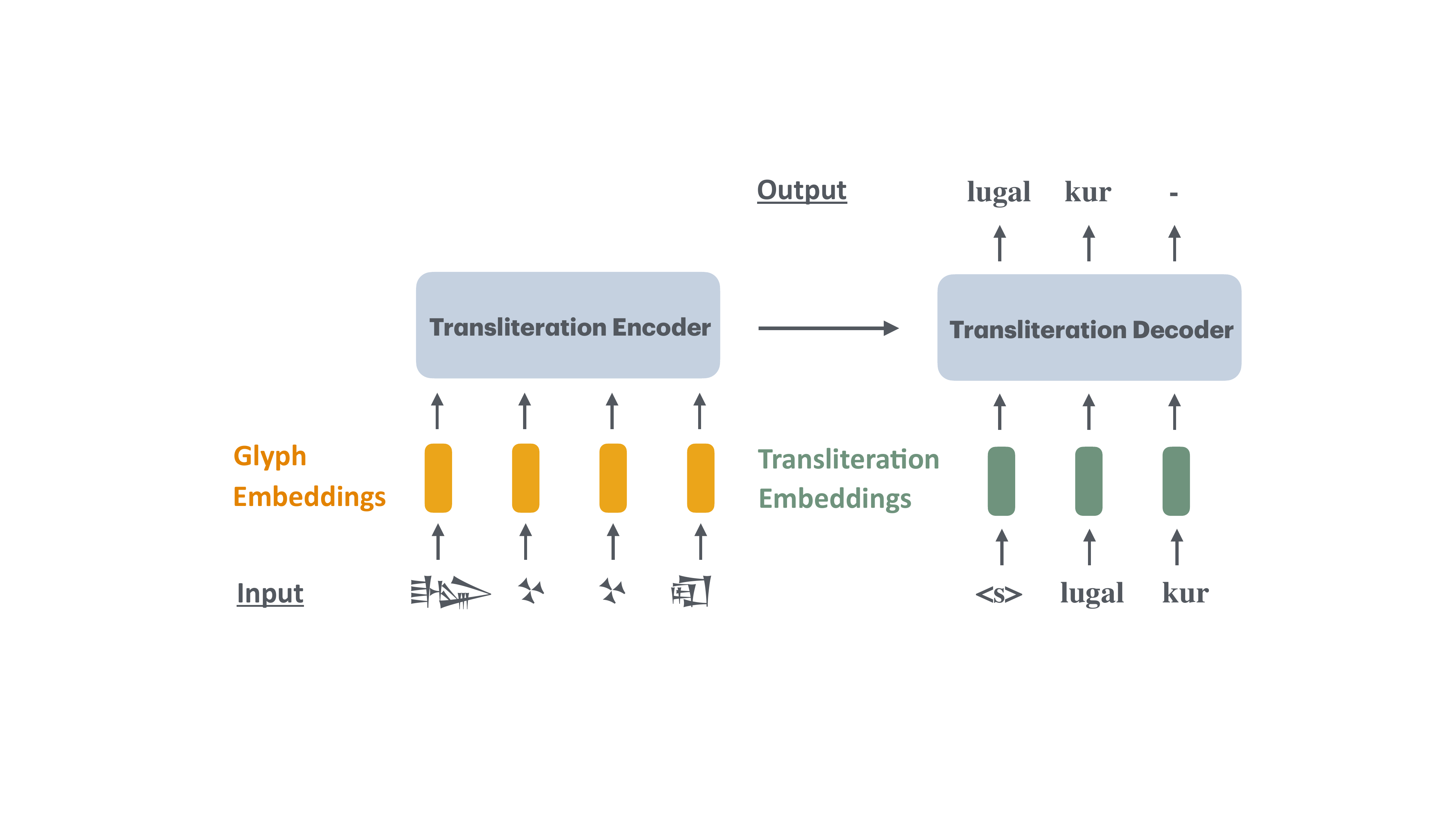}
    \vspace*{-1.7cm}
    \caption{Illustration of the neural baseline model architecture. Inputs are read in as glyph tokens, while outputs are transliteration tokens.}
    \label{fig:model-architecture}
    \vspace*{-0.2cm}
\end{figure}

The lack of tokenization support for Sumerian presents a first challenge in applying the XLM-R model to transliterating Sumerian. To deal with this, we retrain the default SentencePiece tokenizer \citep{Kudo_Richardson_2018} used by the XLM-R model twice: once to build a `glyph tokenizer' that is trained only on the Sumerian glyphs in \textit{SumTablets}, and once to build a `transliteration tokenizer' that is trained only on the corresponding Sumerian transliterations in \textit{SumTablets}. The `glyph tokenizer' has a vocab size of 632 glyph tokens and is used by the encoder model to generate `glyph embeddings' from a string of Unicode-encoded glyphs. The `transliteration tokenizer' has a vocab size of 1024 transliteration tokens and is used by the decoder model to output transliterations. The vocabularies of both the glyph and transliteration tokenizers include eleven special tokens, including the extra-semantic special tokens discussed in section \ref{subsec:extra-semantic-tokens}.

We structure our transliteration model as a sequence-to-sequence (encoder-decoder) model. We initialize both the encoder and decoder separately with the pre-trained weights of an XLM-R model. 

\begin{figure}[ht!]
    \centering
    \vspace*{-1.2cm}
    \hspace*{-0.20\linewidth}
    \includegraphics[width=1.4\linewidth]{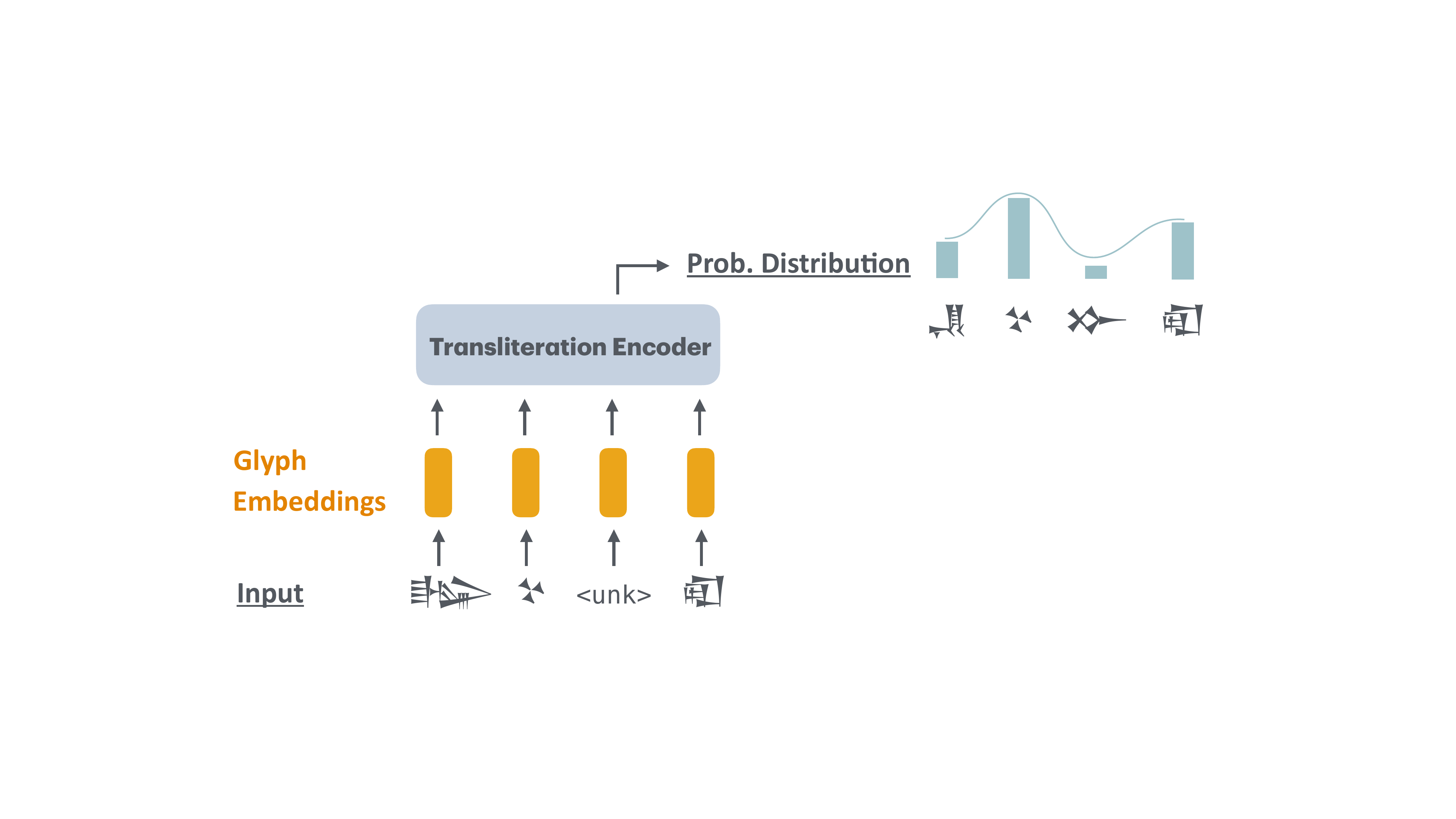}
    \vspace*{-1.9cm}
    \caption{The encoder model can produce a probability distribution over possible glyphs that can replace an {\small <UNK>} token. This is because the encoder is trained using an MLM objective.}
    \label{fig:encoder-model}
    \vspace*{-0.2cm}
\end{figure}

We train the model in three stages: First, we independently fine-tune the pretrained encoder model on the Unicode cuneiform glyphs using a masked language modeling task (MLM). This step yields a model with effective internal representations for the glyphs. Then, we integrate the decoder, training the full encoder-decoder model to take glyph sequences as input and auto-regressively predict target transliterations token-by-token. To stabilize the auto-regressive training of the joint encoder-decoder model, we decompose this process two stages. We first freeze the encoder weights (only training the decoder) for one-third of the time that we train the joint encoder-decoder model. For the rest of training, we unfreeze the encoder weights and allow both the encoder and decoder to receive gradient updates. Figure \ref{fig:model-architecture} showcases the encoder--decoder model architecture. An added benefit of using both an encoder and a decoder is that the encoder can function independently from the decoder to predict missing or unknown glyphs, as illustrated in Figure \ref{fig:encoder-model}.

Both the encoder and decoder are initialized with the pre-trained weights of a 279 million parameter XLM-R model \footnote{For a full description of the XLM-R model, refer to: https://huggingface.co/FacebookAI/xlm-roberta-base}. We initially fine-tune the encoder on the MLM task for 50 epochs, with sequences lengths of 64 tokens, a learning rate of 5e-05, batch size of 2,048, and 200 warmup steps. We set the MLM masking probability to 0.10 and use the same 80-10-10 masking procedure as in \citeauthor{Devlin_EtAl_2019}. Next, the encoder-decoder with frozen encoder weights is trained with a learning rate of 1e-04 for 2 epochs. Finally, we unfreeze the encoder weights and train the full encoder--decoder model with a learning rate of 5e-05 for a further 4 epochs. For both encoder--decoder learning procedures, we set the train batch size to 128 and the number of warmup steps to 100. All training used the AdamW optimizer \citep{Loshchilov_Hutter_2019} and was run on a single A100 SXM 80GB. For transliteration generation, we use beam search decoding with a beam size of 5. 

Throughout our experiments, we set the maximum sequence length to 128. For tablets with more than 128 glyphs, we divide both the pre-tokenized glyphs and transliterations by newlines—these divisions align due to how we design \textit{SumTablets} to preserve tablet structures. We then tokenize chunks of N lines, with N decreasing in size progressively from 16 down to 1, until the resulting chunk contains slightly less than 128 tokens. This segmentation ensures that all resultant chunks contain a maximum amount of tokens within the valid sequence length.

After processing the data into chunks of sequence length 128, we find that the dataset comprises 178,208 administrative examples and 23,282 non-administrative examples. To address the imbalance, we up-sample non-administrative examples by a factor of 5 for the initial two epochs of training and then reduce the up-sampling factor to 3 for the remaining epochs.

% XLM-R Model 
Our encoder-decoder model achieves an average character-level chrF score of 97.54 on unseen test data, setting, to the best of our knowledge, a new state-of-the-art benchmark performance on the transliteration task. We report results for both baselines and across all time periods and genres in Table \ref{tab:results}. Our work demonstrates the capability of large multilingual models to model and transliterate Sumerian, despite the highly fragmented nature of these extant texts and the language being both low-resource and an isolate. 

\subsection{Analysis}
\label{subsection:analysis}

 We derive several key takeaways from our results.

\vspace{0.25cm}

\textbf{The genre of the texts impacts transliteration performance.} The difference in transliteration performance across genres that we observe in Table \ref{tab:results} is intuitive given the nature of the underlying data. Because the training data is dominated by administrative examples, it is natural that that would be the best performing category. These texts also tend to be relatively formulaic. Liturgical, letter, and literary texts, on the other hand, have a different style, form, and vocabulary from the rest of the corpus. These genres (liturgical in particular) are also some of the most challenging for experts. Genre also affects performance insofar as for most genres there are so few examples on which to train or evaluate.

\vspace{0.25cm}

\textbf{Inconsistent transliteration conventions muddle performance.} Some of the different readings for a glyph stem not from a tangible semantic difference but from phonetic or aesthetic disagreement. For instance, ``saŋ'' and ``sag'' represent the same thing, but Assyriologists have a preference in how they represent the nasal `g.' A lack of standardization on matters like this fragments the patterns in which models observe a reading occurring. 

\vspace{0.25cm}

\textbf{It is difficult to predict phonemes in names.} Manual error analysis showed that some errors occurred when selecting a reading that serves as part of a name (playing a phonetic role). Our neural baseline model would often predict a valid reading for a glyph, but a different one than in the true transliteration. Future work will incorporate expert evaluation to determine whether these predictions are any more or less plausible than those in the original transliteration.

% =============================================
% ============== LIMITATIONS ==================
% =============================================
\section{Limitations}
\label{section:limitations}

We note that our work has some limitations, both in terms of the \textit{SumTablets} dataset and the transliteration model. 

\subsection{Dataset Limitations}

Administrative documents have an outsized representation in the train, validation, and test data. This dataset imbalance is a natural by-product of the category of documents produced by Mesopotamian peoples and is an unavoidable consequence of working with Sumerian texts. Although we chose to oversample non-administrative tablets in the train set by a factor of 5 during training of our model, we leave the choice of how to best handle this imbalance to the consumer.

While the set of Unicode cuneiform glyphs is largely complete, there are still glyphs that are not represented in this set, particularly some complex compound glyphs. We currently convert these glyphs and their corresponding readings into {\small <UNK>} tokens, but future work could incorporate unique identifiers for these glyphs as a placeholder until they are added to the Unicode standard.

Finally, there is considerable orthographic variation in glyphs over time, and representing these in Unicode flattens these (potentially meaningful) variations into a single, universal representation.

% * Unicode is not complete
% * Unicode does not reflect orthography

\subsection{Model Limitations}

% todo: break this into two?
In this paper, we train an XLM-R model on \textit{SumTablets} as a fully supervised neural baseline for Sumerian glyph transliteration. We give our model access to the entire training set to explore the limit of a pre-trained cross-lingual model to perform this novel task. Our work, however, does not study the zero- and few-shot abilities of cross-lingual models, which is typically of more interest when evaluating a model's cross-lingual abilities. Nor do we study the performance of a model trained from scratch on our dataset. We encourage future work to use \textit{SumTablets} as a few- and zero-shot cross-lingual benchmark task to evaluate how a multilingual model's language understanding transfers to the Sumerian language.

Moreover, we recognize that the dictionary baseline that we implement is very simple, and that a better point of comparison would be an N-gram model.

% =============================================
% ============== CONCLUSION ===================
% =============================================
\section{Conclusion}

We introduce \textit{SumTablets}, the first collection of paired glyph-transliterations extracted from 91,606 Sumerian tablets. Our dataset provides a resource for experts and non-experts alike to contribute to the development of transliteration models. We define the transliteration task, evaluation method, and establish a baseline performance so that future results may be compared. We also demonstrate that---despite Sumerian's status as a low-resource language and language isolate---large pretrained multilingual language models can be adapted to perform the sequence-to-sequence task of transliterating a sequence of Unicode cuneiform glyphs with remarkable accuracy.

With such an abundance of extant texts and so few specialists capable of reading them, we believe transliteration models will enable Assyriologists to spent less time on tedious, from-scratch transliteration and more time on research and translation.

% x=============================================
% ============== ACKNOWLEDGEMENTS =============
% =============================================
\section*{Acknowledgments}
We would like to thank Niek Veldhuis for his assistance as we explored past digital and computational Assyriological work.

First and foremost we thank the relentless dedication of contributors to the countless open-source cuneiform and Sumerian language resources. We would also like to thank Niek Veldhuis, Sabri Eyuboglu and Geoff Angus for their many suggestions and guidance in creating SumTablets. Richard Diehl Martinez is supported by the Gates Cambridge Trust (grant OPP1144 from the Bill \& Melinda Gates Foundation).

\bibliography{custom}

% \appendix

% \section{Data Packing and Sampling}
% \label{sec:appendix-data-sampling-formatting}

\end{document}